\documentclass[final,3p,times,twocolumn]{elsarticle}

\usepackage{amssymb}
\usepackage{multirow}
\usepackage{rotating}
\usepackage{adjustbox}

\usepackage{hyperref}

\usepackage{mathtools} 
\usepackage{tabularx}

\usepackage{enumitem}
\newlist{todolist}{itemize}{2}
\setlist[todolist]{label=$\square$}
\usepackage{pifont}
\begin{document}

\begin{frontmatter}

\title{Accurate and Fast Fischer-Tropsch Reaction Microkinetics using PINNs}

\author[stcb]{Harshil Patel\corref{cor1}}
\ead{Harshil.Patel@shell.com}

\author[stcb]{Aniruddha Panda\corref{cor1}}
\ead{Aniruddha.Panda@shell.com}

\author[softs,tsnu]{Tymofii Nikolaienko}

\author[stca]{Stanislav Jaso}

\author[stca]{Alejandro Lopez}

\author[stcb]{Kaushic Kalyanaraman}

\cortext[cor1]{Both authors contributed equally, corresponding authors}

\affiliation[stcb]{organization={Shell India Markets Pvt. Ltd.},
            addressline={Mahadeva Kodigehalli}, 
            city={Bengaluru},
            postcode={562149}, 
            state={Karnataka},
            country={India}
            }

\affiliation[stca]{organization={Shell Global Solutions International B.V.},
            addressline={P.O. Box 38000}, 
            city={Amsterdam},
            postcode={1030 BN}, 
            country={The Netherlands}
            }

\affiliation[softs]{organization={ SoftServe, Inc.}, 
            addressline={2d Sadova Str.}, 
            city={Lviv},
            postcode={79021}, 
            country={Ukraine}
            }

\affiliation[tsnu]{organization={Faculty of Physics of Taras Shevchenko National University of Kyiv},
            addressline={64/13 Volodymyrska Str.}, 
            city={Kyiv},
            postcode={01601}, 
            country={Ukraine}
            }

\begin{abstract}
Microkinetics allows detailed modelling of chemical transformations occurring in many industrially relevant reactions. Traditional way of solving the microkinetics model for Fischer-Tropsch synthesis (FTS) becomes inefficient when it comes to more advanced real-time applications. In this work, we address these challenges by using physics-informed neural networks(PINNs) for modelling FTS microkinetics. We propose a computationally efficient and accurate method, enabling the ultra-fast solution of the existing microkinetics models in realistic process conditions. The proposed PINN model computes the fraction of vacant catalytic sites, a key quantity in FTS microkinetics, with median relative error (MRE) of 0.03\%, and the FTS product formation rates with MRE of 0.1\%. Compared to conventional equation solvers, the model achieves up to $10^6$ times speed-up when running on GPUs, thus being fast enough for multi-scale and multi-physics reactor modelling and enabling its applications in real-time process control \& optimization.
\end{abstract}




\end{frontmatter}


\newpage

\section{Introduction}
\label{sec:intro}

Accurate modelling of chemical reaction kinetics is vital to understand chemical processes and to predict their behaviour \cite{Okino-simplification-1998, van-Riel-dynamic-2006, Green-moving-2020}. It relies on expressing the chemical reaction mechanisms in a mathematical form \cite{Green-moving-2020} for applications such as optimization of the properties of the catalysts \cite{Matera-progress-2019, Notagamwala-microkinetic-2021}, selecting proper design and operating conditions for chemical reactors \cite{Coker-modeling-2001, Labovsky-mathematical-2007}, enabling real-time process optimization \cite{Wentrup-dynamic-2022} etc. In complex industrial chemical plants \cite{Wentrup-dynamic-2022}, kinetics of the involved chemical transformations can be considered as the core building block \cite{Marin-kinetics-2021}, which is then used to create larger-scale models \cite{Moser-mechanistic-2021, Ho-kinetic-2008}, e.g., capable of being used as a digital twin \cite{Tao-make-2019, Tao-digital-2019, Gargalo-towards-2021} of the chemical plant \cite{Nandakumar-perspectives-2022}. From such ground-up perspective, accuracy and  reliability  \cite{Ozbuyukkaya-determining-2022} becomes one of the key requirements for a kinetic model \cite{Matera-progress-2019}.

One way to meet these requirements is by using a first-principle approach to the reaction kinetics, in which the ultimate reactants-to-products transformation is decomposed into a set of elementary reaction steps and  all intermediates are tracked using the chemical insights about the underlying mechanisms \cite{Green-moving-2020, Marin-kinetics-2021, Susnow-rate-based-1997}. Each elementary step \cite{Kaufman-rates-1985} is then characterized by the thermodynamic and kinetic properties \cite{Pollak-reaction-2005, Henriksen-theories-2018}. These properties are then used as parameters in equations to relate the rates of change in the concentrations of the constituent species \cite{Carr-chapter-2007, Santiso-multi-scale-2004}. The whole approach is often referred to as microkinetics modelling, when the chemical transformation of interest occurs at the surface of a catalyst as a 'microdomain' in which the reactions take place and the role of adsorption is taken into account explicitly \cite{Notagamwala-microkinetic-2021, Marin-kinetics-2021, Bartholomew-fundamentals-2011, Dumesic-microkinetics-1993}. Due to the explicit consideration of the role of the surface processes in this approach, it typically involves an additional stage of solving a set of balance equations to find the surface coverage of all adsorbed species \cite{Notagamwala-microkinetic-2021}.

The benefits of microkinetics approach come at the cost of its complexity. However, overcoming this barrier by dedicating research efforts becomes justified when the chemical process of interest has a potential of having high economic and/or environmental impact.  One such example is the chemical transformation of the carbon molecules present in crude oil, natural gas (including the associated natural gas produced from oil wells as a byproduct \cite{Zhang-Efficient-2016}), shale gas, biogas, coal, biomass or waste \cite{Centi-Chemistry-2020, Paladino-closed-2018} into usable fuels (typically, short chain carbon molecules) or other valuable chemicals (e.g., methanol, dimethyl ether, dimethyl carbonate etc \cite{Garciagarcia-analytical-2021, Masudi-recent-2020, Lee-liquid-2005}). Utilization of alternative carbon sources is considered as an essential step in maintaining sustainable production of chemical building blocks for much of the chemical industry \cite{Graves-sustainable-2011}. In cases when the ultimate product of the chemical transformation is liquid fuels, the processes can be unified under the umbrella term XTL (X-to-Liquid) \cite{de-Klerk-overview-2011, Martinelli-overview-2020, Sonal-recent-2021} where, X can be anything depending on the carbon feed stock used \cite{Cheng-advances-2017, de-Klerk-10-2020}, e.g., CTL (coal to liquids) \cite{Hook-review-2010, Jin-indirect-2014}, GTL (gas to liquids) \cite{Wood-gas-liquids-2012, Sonal-recent-2021}, and BTL (biomass to liquids) \cite{Swain-biomass-2011, Lappas-18-2016}.

Carbon from atmosphere or industry-generated flue gases in the form of carbon dioxide, $\mathrm{CO_2}$, can be recycled and further used as a resource as well, by one of the processes collectively known as Carbon capture and utilization (CCU) techniques \cite{Garciagarcia-analytical-2021}. When used in conjunction with renewable energy sources \cite{Graves-sustainable-2011}, CCU can reduce the environmental impacts of manufacturing processes producing valuable chemicals, e.g., reduce $\mathrm{CO_2}$ emissions and the dependence on fossil fuels as carbon source.
Catalyst-dependent conversion of $\mathrm{CO_2}$ into methane \cite{Aziz-co2-2015, Younas-recent-2016} can also be used for power-to-gas \cite{Frontera-Supported-2017} (and, generally, power-to-X \cite{Centi-Chemistry-2020, Bailera-review-2021}) conversion in renewable energy storage systems \cite{Pearson-energy-2012}, essential for stabilizing electrical grids powered by sources with discontinuous production like photovoltaics and wind turbines\cite{Centi-Chemistry-2020}.
Another aspect of CCU is its relevance for the production of carbon-neutral fuels \cite{Graves-sustainable-2011, France-chapter-CO2-2015}, which have net zero $\mathrm{CO_2}$ footprint \cite{Pearson-energy-2012, Zeman-carbon-2008} or greenhouse gas emissions.
Various means of attaining carbon-neutral hydrocarbons have been reviewed in \cite{Zeman-carbon-2008, Davis-net-zero-2018}, as well as the implementation aspects \cite{Chen-strategies-2022} and challenges \cite{van-der-Giesen-energy-2014} of achieving carbon neutrality strategies in practice.
In various carbon utilization pathways \cite{Hepburn-technological-2019}, including XTL and CCU, a mixture of carbon monoxide ($\mathrm{CO}$) and hydrogen ($\mathrm{H}_2$) gas known as syngas, acts as the key intermediate stream \cite{Centi-Chemistry-2020}, which is then used to produce hydrocarbon fuels. Such a two-stage process of producing fuels from a carbon-containing feed appears to be more practical and technologically mature \cite{Cheng-advances-2017, Macdowell-overview-2010} as compared to the direct conversion.

Fischer-Tropsch synthesis (FTS) process is one of the most applied techniques for catalytic conversion  of syngas into fuels in the form of long-chain hydrocarbons \cite{Centi-Chemistry-2020, Martinelli-overview-2020, Sonal-recent-2021, Mendez-kinetic-2020, van-der-Laan-kinetics-1999, Martin-chapter-2016}. FTS dates its history back to 1926 \cite{France-chapter-CO2-2015} (see \cite{Steynberg-chapter-2004, Schulz-short-1999} for historical overview) and has been implemented in a variety of industrial setups since then (see \cite{TodicThesis2015} for illustrative examples). 
FTS process relies on the use of a catalyst, typically based on iron or cobalt and/or their oxides loaded onto a porous substrate in the form of pellets, to convert the syngas into hydrocarbons.
The actual composition of products which are formed as a result of the FTS process varies depending on the operating conditions. It majorly consists of a mixture of n-paraffins (linear 1-alkanes) and 1-olefins (alkene containing a single carbon–carbon double bond in the terminal position) with number of carbon atoms from 1-4 (gases), through 5-20 (liquids) and higher (waxes) \cite{Wentrup-dynamic-2022, Cheng-advances-2017, Mendez-kinetic-2020, Martin-chapter-2016}. 
The liquid fraction of FTS products is highly-valuable as ultra-clean liquid fuels \cite{TodicThesis2015} due to it being almost free of aromatic compounds, nitrogen, sulfur,  and other toxic substances typically found in petroleum products and are known to violate strict environmental requirements \cite{Cheng-advances-2017, Steynberg-chapter-2004}.  Another advantage of the synthetic fuels is their usability within the current petroleum refining and distribution infrastructure \cite{Zeman-carbon-2008, Graves-sustainable-2011}.
Products with carbon numbers (number of carbon atoms in a hydrocarbon) 5-11 can be used to produce gasoline, whereas the synthesis products with carbon numbers in the 10-20 range can be used to produce diesel (“green diesel”) \cite{Mendez-kinetic-2020, Teimouri-kinetics-2021}.Therefore, while optimizing the reactor operating conditions, achieving high selectivity towards the desired products is as important as maximizing the overall conversion \cite{France-chapter-CO2-2015}. Detailed microkinetics is thus an essential ingredient in the robust optimization of the FTS reactor performance.

The sequence of elementary transformations, collectively referred to as the reaction mechanism of the FTS process received substantial attention due to its key importance as the basis for building the microkinetics model of the process, and has been covered by several reviews \cite{Martinelli-overview-2020, Sonal-recent-2021, Mendez-kinetic-2020, van-der-Laan-kinetics-1999, Teimouri-kinetics-2021, Mahmoudi-review-2017, Dry-practical-1996, Davis-fischertropsch-2001, Adesina-hydrocarbon-1996}. These mechanisms are generally grouped into alkyl, CO insertion, enolic, and alkenyl types, while the relative importance of different mechanisms depends on the specific catalyst and process conditions (e.g., $\mathrm{H_2}/\mathrm{CO}$ ratio, the presence of other substances). Accordingly, several microkinetics models exist in literature. In particular, in \cite{Todic-CO-insertion-2014, Todic-corrigendum-2015}, the authors present a detailed kinetic model (reviewed below in this paper) of the FTS process based on the CO-insertion mechanism. This model accounted for the dependence of 1-oleﬁns desorption on the length of carbon chain and the distribution of synthesis products obtained from the proposed model was shown to agree with the experimental data.

The improvements in the description of the experimental data offered by the aforementioned model came at the cost of making the overall description computationally more expensive. Specifically, a set of non-linear highly-coupled algebraic equations describing the chain growth probability and the fraction of vacant catalytic sites need to be solved for each new instance of the model inputs, viz., the temperature and partial pressures of $\mathrm{H_2}$, $\mathrm{CO}$ and $\mathrm{H_2 O}$. Unless implemented in a highly-parallelized manner, solving the set of non-linear equations can become a computational bottleneck, especially in use-cases where the microkinetics model is a component of a more complex reactor model that is designed to optimize its operational parameters. Due to this computational bottleneck, the calculation of the derivatives of the reaction rates with respect to the input parameters can become even more challenging, although such gradients are usually desirable for using the model in any optimization scenarios.

Machine-learning (ML) methods are an attractive alternative when it comes to building automatic and versatile approximators for complex dependencies. 
In particular, neural networks (NNs) are known for their ability to approximate complex and highly non-linear relationships by minimzing a prescribed loss function and thus, automatically 'learning' them from the collection of pre-computed 'ground truth' input-output examples that is often referred to as training data.
At the same time, the high flexibility of NNs as universal function approximators requires a large amount of training data to generalize well.
In applications, where obtaining each new data point requires an additional run of a stand-alone simulation program, e.g., a conventional equation solver, generating a sufficiently large number of data points for training becomes a time-consuming task or even practically impossible.
This problem becomes more evident in cases where the solution of ordinary differential equations (ODEs) or partial differential equations (PDEs) need to be approximated using a NN. Such interdependent set of equations are a key ingredient of multiscale and multiphysics problems in scientific and engineering applications, where the conventional method of solving them using numerical methods is often computationally expensive \cite{Pearce-exploring-2019, Babur-survey-2015, Montero-chacon-computational-2019}.

Scientific machine learning or Physics-informed machine learning \cite{Raissi-physics-informed-2019, Karniadakis-physics-informed-2021, Wang-learning-2021, Cuomo-scientific-2022} is a relatively new approach for building and training ML models in agreement with some additional knowledge, expressible in the form of equations, enabling creation of hybrid physics-based and data-driven models \cite{Wang-hybrid-2022}.
Among the different ways of incorporating additional knowledge into ML models \cite{Karniadakis-physics-informed-2021, Bronstein-geometric-2017}, physics-informed neural networks (PINNs) stand out as an approach allowing the use of NNs to approximate the solution of equations, e.g., ODEs or PDEs, without the need of having pre-computed training dataset.
This approach is thus well suited for cases when traditional solvers for the equations of interest are computationally expensive.
In physics-informed training the objective is to minimize the defined loss function so as to measures how well the dependency learned by NN satisfies the given equations. This is in contrast to a traditional regression-type training set-up where the loss function measures the discrepancy between the ground-truth outcome and the prediction made by the NN.
Apart from being suitable for solving forward and inverse problems \cite{Raissi-physics-informed-2019, Karniadakis-physics-informed-2021, Hennigh-NVIDIA-2021}, the PINNs approach brings an additional important benefit over the conventional solvers due to its ability of solving parameterized problems  \cite{Karniadakis-physics-informed-2021, Cuomo-scientific-2022}, i.e., training a single NN capable of approximating not only a single solution to the given equation, but a set of parameterized solutions at once.
Due to this, the effective speed-up obtained during inference with PINN becomes even more favorable as compared to the conventional equation solvers.

In this work, we show how PINN can be used to solve a set of non-linear and highly coupled algebraic equations describing the FTS process while also achieving significant computational speedup at almost the same accuracy in comparison with the conventional solver. The proposed PINN model computes the fraction of vacant catalytic sites and the chain growth probabilities as functions of $\mathrm{H}_2$, $\mathrm{CO}$ and $\mathrm{H_2 O}$ partial pressures and the temperature in the reacting system. These quantities are then used to compute the production/reaction rates of n-paraffin and 1-olefins.
Our approach is fast enough to be considered as an on-the-fly solver for more complex optimization and control systems. We further discuss the implementation details of PINN, present it's training process and discuss the speed-up and accuracy benchmarking results.

In the rest of the paper we first briefly review the set of closely related FTS microkinetics models, which are referred to as Todic's class of models, and discuss the type of equations which are needed to be solved in this approach. 
Then, we introduce the PINNs approach to solve such a system of equations followed by the PINN training procedure and the analysis of the performance of the trained model in terms of accuracy and speed.
We close the discussion with an outlook of some further directions.

\section{Microkinetic modelling and PINNs approach}

\subsection{Kinetic model of FTS process}

\newcommand{\GeneralPapersOnFTSmechanisms}{
\cite{Sonal-recent-2021, Mendez-kinetic-2020, van-der-Laan-kinetics-1999}
}

FTS process is generally considered as a polymerization type reaction \GeneralPapersOnFTSmechanisms when building a microkinetics model. It starts with absorbing the $\mathrm{CO}$ and $\mathrm{H_2}$ reactants on the surface of the catalyst (absorption step), proceeds through the steps of chain initiation (when an initial substance is formed on the surface and acts as a `seed' from which the new hydrocarbon chain can then grow), chain propagation (when new carbon atoms are added to the chain thus increasing its length), and ends with chain growth termination step. 
The process is finalized by the desorption of the synthesized molecule from the surface of the catalyst.
Ideally, each of these steps should be described by a series of elementary chemical reactions, accounting for all the possible intermediates involved in the transformations.
Moreover, some additional steps can be considered \cite{Yang-detailed-2003}, e.g., re-adsorption of the formed products onto the surface of the catalyst or their involvement into some secondary reactions.
However, due to the complexity of the entire process, description of its steps is thus usually simplified to a certain degree. 

The Anderson–Schulz–Flory (ASF) model \cite{Flory-molecular-1936, Schulz-uber-1939, Anderson-fischertropsch-1951, Anderson-catalysts-1956} (see \cite{Anderson-schulz-flory-1978} for a historical perspective), which focuses on the chain growth step, is one of most basic approaches to obtain the distribution of FTS products.
At each instance, the growing chain in this model can either get elongated by a new $-\mathrm{CH_2}-$ group with probability $\alpha$ (which is thus called the chain growth probability), or otherwise the chain growth is terminated with probability $1-\alpha$ and the formed chain is detached from the surface.
That leads to the probability $w_n = \alpha ^{n-1} \cdot (1 - \alpha)$ of obtaining the product with $n$ carbon atoms \cite{Sonal-recent-2021, Cheng-advances-2017}. Despite being intuitively appealing and relying on a single parameter $\alpha$, the ASF model yields the distribution of FTS products which is only in qualitative, but not quantitative agreement with the experiments. To build the model suitable for making quantitative predictions, the chain growth probability $\alpha$ needs to be made product-dependent (see \cite{Kruit-selectivity-2013, Matsuka-effect-2016, Song-effect-2006} for the discussion of different factors affecting $\alpha$). This is done in non-ASF models.

Within the microkinetics approach, the the chain growth probability is related to a particular product by considering the elementary chemical reactions occurring at each step of the FTS process. 
Several models for the exact sequence of these reactions have been proposed, each one assuming its own set of intermediate substances as being the most relevant for the particular catalyst of interest, thus introducing the corresponding overall chain growth mechanism (e.g., alkyl, enolic, CO insertion, alkenyl) \GeneralPapersOnFTSmechanisms.

Herein, we focus on the model based on the CO-insertion mechanism \cite{Todic-CO-insertion-2014}, but technically, the PINNs approach we propose in this work is also applicable to a much wider class of models, which we refer to as Todic's class of models.
These models have some differences in terms of underlying set of elementary chemical reactions, but have two key similarities. One of them is the assumption that because of the action of the weak Van der Waal's interactions between the 1-olefin precursor and the surface of the catalyst, the activation energy of the 1-olefin desorption $E_{d,o}^{n}$ increases linearly with the number of carbon atoms $n$ in the molecule \cite{TodicThesis2015, Todic-CO-insertion-2014, Todic-optimization-kinetic-2013}, viz., 
$E_{d,o}^{n} = E_{d,o}^{0} + n \cdot \Delta E$.
In turn, that implies that the rate constant for 1-olefins desorption depends on $n$ exponentially (${\sim} e^{c \cdot n}$, where $c = -\frac{\Delta E}{R T}$). Secondly, the models within Todic's class are similar in considering a state referred to as a growing chain intermediate \cite{Todic-optimization-kinetic-2013} and denoted by $\mathrm{C}_n \mathrm{H}_{2 n + 1} - \mathrm{S}$. 
The way in which this state is formed is model-dependent, but once it is achieved, it can only then either undergo an elementary reaction yielding the finalized hydrocarbon chain (a paraffin or 1-olefin), or be consumed by the chain of elementary reactions comprising the chain propagation step of FTS and ending with $\mathrm{C}_{n+1} \mathrm{H}_{2 (n+1) + 1}-\mathrm{S}$ state. 
Schematically, these alternatives can be denoted as 
$$
\mathrm{C}_n \mathrm{H}_{2 n + 1} - \mathrm{S} 
\left\{
  \begin{array}{l}
    \xrightarrow{k_{growth}} \ldots \xrightarrow{} \mathrm{C}_{n + 1}
    \mathrm{H}_{2 (n + 1) + 1} - \mathrm{S}\\
    \xrightarrow{k_{paraffin,n}} \mathrm{C}_n \mathrm{H}_{2 n + 2}\\
    \xrightarrow{k_{olefin,n}} \mathrm{C}_n \mathrm{H}_{2 n}
  \end{array}
\right.
$$
where, $\mathrm{C}_n \mathrm{H}_{2 n}$ option is present for $n \ge 2$ only, `$\ldots$' denotes the reactions comprising the chain propagation step and $k_{growth}$, $k_{paraffin,n}$, $k_{olefin,n}$ are the reaction rate constants of the appropriate elementary steps in the FTS model (e.g., \cite{Todic-optimization-kinetic-2013}).
The chain growth probability $\alpha_n$ is then introduced as the ratio \cite{Sonal-recent-2021, Mendez-kinetic-2020, van-der-Laan-kinetics-1999}
$$
\alpha_n = \frac{k_{growth}}{k_{growth} + k_{paraffin,n} + k_{olefin,n} }
$$
where $k_{olefin,n} = 0$ for $n=1$ (when no 1-olefin can be formed).
Dependence of $\alpha_n$ on the number of carbon atoms $n$ appears because $k_{olefin,n}$ changes with $n$ as $\sim e^{c \cdot n}$, while $k_{paraffin,n}$ is independent on $n$ when $n \ge 2$, but can have a different value when $n=1$ (allowing for a special treatment of $\mathrm{CH_4}$ as a FTS product).

In case of a stationary process, which is usually considered in the Todic's class of models, $\alpha_n$ can be shown to be equal to the ratio of equilibrium surface coverages \cite{Todic-optimization-kinetic-2013}, $
\alpha_n = \frac{ [\mathrm{C}_n \mathrm{H}_{2 n + 1} - \mathrm{S}] }{ [\mathrm{C}_{n-1} \mathrm{H}_{2 (n-1) + 1} - \mathrm{S}] }
$ (when $n \ge 2$) or  $
\alpha_1 = \frac{ [\mathrm{C} \mathrm{H}_3 - \mathrm{S}] }{ [\mathrm{H} - \mathrm{S}] } 
$ (when $n=1$).

In addition to the surface coverage of $\mathrm{C}_n \mathrm{H}_{2 n + 1} - \mathrm{S}$, the equilibrium fractions of catalyst surface sites occupied with all other possible reactions intermediates can be obtained, and then related to the fraction of vacant catalytic sites $\mathrm{[S]}$ by the expressions depending on $\alpha_n$. 
That leads to the so-called site balance equation which has the same form in the entire Todic's class of FTS microkinetics models \cite{Todic-CO-insertion-2014, Todic-corrigendum-2015, Todic-optimization-kinetic-2013}. It reads
\begin{equation}
 \label{eq:S-equation}
[S]
=
\frac{1}{
c_0 + c_{S} \cdot
  \left(\alpha_1  + \alpha_1 \alpha_2 +  \alpha_1 \alpha_2 \sum\limits_{i=3}^{N_{max}} \prod\limits_{j=3}^i \alpha_j\right)
  } ,
\end{equation}
where $N_{max}$ is the largest possible number of carbon atoms in FTS product molecule.
By solving non-linear equation \ref{eq:S-equation}, coupled with the expressions relating $\alpha_n$ to $[S]$, the fraction of vacant sites $[S]$ can be found and then used to obtain detailed information about the products of FTS and their formation rates.
The coefficients $c_0$ and $c_{S}$ in equation \ref{eq:S-equation} are determined by the reaction conditions and their explicit expressions are model-dependent.

Hereinafter our consideration will be limited to the specific case of $\mathrm{CO}$ insertion mechanism microkinetics model, as proposed in \cite{Todic-CO-insertion-2014, Todic-corrigendum-2015}. In that instance of Todic's class of models, the following expressions are used
\begin{equation}
 \label{eq:c_0}
c_0 = 1 + K_1 P_{\mathrm{CO}} + \sqrt{K_2 P_{\mathrm{H}_2}} \; ,
\end{equation} 
\begin{equation}
 \label{eq:c_S}
c_{S} = \frac{1}{K_2^2 K_4 K_5 K_6} \frac{P_{\mathrm{H}_2 \mathrm{O}}}{P_{\mathrm{H}_2}^2}+\sqrt{K_2 P_{\mathrm{H}_2}}  \; ,
\end{equation} 
while the explicit expressions for $\alpha_n$ in  equation \ref{eq:S-equation} are
\begin{equation}
 \label{eq:alpha_1}
\alpha_1 = \frac{k_3 K_1 P_{\mathrm{CO}}}{k_3 K_1 P_{\mathrm{CO}}+k_{7 \mathrm{M}} \sqrt{K_2 P_{\mathrm{H}_2}}}  \; ,
\end{equation} 
\begin{equation}
 \label{eq:alpha_2}
\alpha_2 = \frac{k_3 K_1 P_{\mathrm{CO}}[S]}{k_3 K_1 P_{\mathrm{CO}}[S]+k_7 \sqrt{K_2 P_{\mathrm{H}_2}}[S]+k_{8, E} e^{c \cdot 2}} 
\end{equation} 
and
\begin{equation}
 \label{eq:alpha_3n}
\alpha_n = \frac{k_3 K_1 P_{\mathrm{CO}}[S]}{k_3 K_1 P_{\mathrm{CO}}[S]+k_7 \sqrt{K_2 P_{\mathrm{H}_2}}[S]+k_{8,0} e^{c \cdot n}} 
\end{equation} 
when $n\ge 3$.
Reaction rates for the final products of FTS are then found as
\begin{equation}
 \label{eq:R_CH4}
R_{\mathrm{CH}_4}=k_{7 \mathrm{M}} K_2^{0.5} P_{\mathrm{H}_2}^{0.5} \alpha_1 \cdot[\mathrm{S}]^2
\; ,
\end{equation}
\begin{equation}
 \label{eq:R_CnH2np2}
R_{\mathrm{C}_n \mathrm{H}_{2 n+2}}=k_7 K_2^{0.5} P_{\mathrm{H}_2}^{0.5} \alpha_1 \alpha_2 \prod_{i=3}^n \alpha_i \cdot[\mathrm{S}]^2
\end{equation}
when $n \geq 2$,
\begin{equation}
 \label{eq:R_C2H4}
R_{\mathrm{C}_2 \mathrm{H}_4}=k_{8 E, 0} e^{c \cdot 2} \alpha_1 \alpha_2 \cdot[\mathrm{S}]
\; ,
\end{equation}
\begin{equation}
 \label{eq:R_C2H2n}
R_{\mathrm{C}_n \mathrm{H}_{2 n}}=k_{8,0} e^{c \cdot n} \alpha_1 \alpha_2 \prod_{i=3}^n \alpha_i \cdot[\mathrm{S}]
\end{equation}
when $n \geq 3$.
In these equations, 
$P_s$ denotes the partial pressure (in MPa) of  substance $s$ ($s = \mathrm{H_2}, \mathrm{CO}, \mathrm{H_2 O}$), 
$K_j = A_j \cdot \exp\left( - \frac{\Delta H _ j}{R T} \right)$ are equilibrium constants for $j$-th elementary step,  
$k_i = A_i \cdot \exp\left( - \frac{E _ i}{R T} \right)$ are reaction rate constant for elementary step $i$, 
$T$ is temperature (in K), 
$  c=-\frac{\Delta E}{R \, T} $ is the factor expressing dependence of effective reaction rates on the number of carbon atoms in the olefin chain.

Here, $A_j$, $\Delta H_j$, $E_i$ and $\Delta E$ are  adjustable parameters of the model.
Their numerical values are usually found by fitting the model predictions (mostly, related to the distribution of FTS products) to the experimental data.
From that perspective, and because the chain of elementary transformations used in the model is never exhaustive \cite{Bukur-role-2016, Yang-detailed-2003}, we prefer to consider the entire set of model's expressions primarily as an interpolation formulae, backed by certain chemical knowledge, rather than as a fully first-principle ground-up approach \cite{Green-moving-2020}.
Such treatment also validates the compensation of some imperfections of underlying theoretical description by tolerating slight shifts of numerical values of the model parameters away from their `microscopically correct' values during the fitting.
Interpolation-based treatment thus allows for widest possible applicability of the model in terms of reaction conditions ranges, which is essential for engineering workflows.
Still, it is the accounting for the dependence of 1-oleﬁns desorption on the carbon chain length which be considered as a main qualitative reason allowing for such a degree of model's flexibility that it can fit distribution of FTS products in good agreement with experimental data.

Since the equations \ref{eq:S-equation}--\ref{eq:alpha_3n} are non-linear with respect to $[S]$ and the r.h.s. of equation \ref{eq:S-equation} has sum-product of $N_{max}$ terms (which is usually quite large, $\approx 10^2$), it needs to be solved numerically, comprising the most computationally expensive step in the entire model.
Technically, this can be done by one of the many well-established root finding techniques which are usually iterative in nature. Alternative is the PINN approach which we present in more details in the next subsection.

When equations \ref{eq:S-equation}--\ref{eq:alpha_3n} are solved with a particular method and $[S]$ and $\alpha_n$ are found, the production rates for all FTS products can be obtained using equations \ref{eq:R_CH4}--\ref{eq:R_C2H2n}.
These rates can then be used, e.g., to optimize the reaction conditions or as the quantities which should be fitted to the experimental data.
In both cases, it is desirable to have the ability to easily compute the derivatives of the reaction rates with respect to reaction conditions, viz., $P_{\mathrm{CO}}$, $P_{\mathrm{H_2}}$, $P_{\mathrm{H_2 O}}$ and $P_{T}$.
This can be achieved easily in PINN approach as well, although can be challenging with traditional equation solvers.

\subsection{PINNs approach for microkinetics modelling}

In the PINNs approach, instead of solving equations \ref{eq:S-equation}--\ref{eq:alpha_3n} the goal is to build a function approximator in the form of a neural network based on these explicit expressions, which would take a four-component row-vector of reaction conditions 
$X = \left\{ 
p_{\mathrm{CO}}, p_{\mathrm{H}_2}, p_{\mathrm{H}_2 \mathrm{O}}, T
\right\}$
as its argument and calculate the fraction of vacant catalytic sites $[S]$, satisfying equations \ref{eq:S-equation}--\ref{eq:alpha_3n}.
We build this function by combining a predefined analytical expressions with a NN performing the most complicated part of input-to-output mapping.

At the beginning, we non-dimensionalize the inputs by using min-max technique and introduce a new vector with components
$$
\bar{X}_j = \frac{X_j - X_j^{min}}{X_j^{max} - X_j^{min}}
$$
where $X_j^{min}$ and $X_j^{max}$ define the range in physical units, within which the $j$-th input variable can change. This transformation facilitates NN training because each non-dimensionalized variable $\bar{X}_j$ always lies in a predefined range from 0 to 1 only. Because the first three components of $X$ are pressures and we put $p_{\mathrm{CO}}^{\min} = p_{\mathrm{H}_2}^{\min} = p_{\mathrm{H}_2 \mathrm{O}}^{\min} = 0$, leading to
\begin{equation}
  \label{eq:X_to_Xbar}
\bar{X} = \left\{ \frac{p_{\mathrm{CO}}}{p_{\mathrm{CO}}^{\max}},
\frac{p_{\mathrm{H}_2}}{p_{\mathrm{H}_2}^{\max}}, \frac{p_{\mathrm{H}_2
\mathrm{O}}}{p_{\mathrm{H}_2 \mathrm{O}}^{\max}}, \frac{T - T^{\min}}{T^{\max}
- T^{\min}} \right\} \; . 
\end{equation}
Here, $p_{\mathrm{CO}}^{\max} = p_{\mathrm{H}_2}^{\max} = 6 \; \mathrm{MPa}$, $p_{\mathrm{H}_2 \mathrm{O}}^{\max} = 1 \; \mathrm{MPa}$, $T^{\min} = 473.15 \textrm{ K}$ and $T^{\max} = 513.15 \textrm{ K}$ and are determined by the typical ranges of reaction conditions anticipated in a real-world FTS reactor.
The final $[S]$ is then computed as 
\begin{equation}
  \label{eq:S_eq_10_Sbar}
  [S] = 10^{- \bar{S} (\bar{X}) } \; , 
\end{equation}
where $\bar{S} (\bar{X}) $ is a function approximated by a NN.

In the proposed implementation, we represent $\bar{S} (\bar{X})$ by a feed-forward fully-connected neural network. That is, $\bar{S} (\bar{X})$ is built as a composition
\begin{equation}
  \label{eq:NN_as_eq}
  \bar{S} (\bar{X}) = l_d (l_{d - 1} (\ldots l_1 (\bar{X}))) \cdot W^{d + 1} + b^{d + 1}
\end{equation}
of `layers' $l_i (x)$, $i = 1, \ldots, d$, which are vector-valued functions  of vector argument $x$.
Each layer is given by $l_i (x) = \sigma (x \cdot W^i + b^i)$ and performs two subsequent transformation.
First, it linearly transforms the argument $x$ into a new vector $L_i = x \cdot W^i + b^i$ (where `$\cdot$' denotes matrix multiplication and $x$ is considered to be a row vector), using the weights matrix $W^i$ and the bias vector $b^i$.
Then, it applies some non-linear scalar `activation function' $\sigma$ to $L_i$ in a component-wise manner to produce the layer's output. 
All components of the weights matrices and biases vectors are considered as adjustable parameters and are computed during NN training.

There are two approaches to train the aforementioned NN.
One of them is `data-driven' or regression-based. In this approach, the weights and biases are computed so as to minimize the discrepancy between the outputs of NN on the set of inputs with the known `ground-truth' outputs. In our case, training NN by this method would clearly require first solving multiple instances of equations \ref{eq:S-equation}--\ref{eq:alpha_3n} with some external solver, to generate sufficient number of input-output pairs for training. Instead, we follow another approach, described in the next subsection, known as the physics-informed method of NN training and thus refer the ultimate network as PINN.

After the PINN is trained, the fraction of vacant catalytic sites $[S]$ can be evaluated for any given combination of input parameters 
$p_{\mathrm{CO}}$, $p_{\mathrm{H}_2}$, $p_{\mathrm{H}_2 \mathrm{O}}$, $T$ within the ranges used during the training using equations \ref{eq:X_to_Xbar}, \ref{eq:NN_as_eq}, \ref{eq:S_eq_10_Sbar}.
Importantly, the derivatives of $[S]$  with respect to any of these inputs can be evaluated analytically. 
In practice, this can be done using the automatic differentiation capabilities \cite{Baydin-automatic-2017, Ketkar-automatic-2021, van-Merrienboer-automatic-2018} offered by virtually all modern deep learning frameworks \cite{Abadi-tensorflow-2016, Paszke-pytorch-2019}.

\subsection{PINN training process}

\begin{figure*}
    \includegraphics[width=\textwidth]{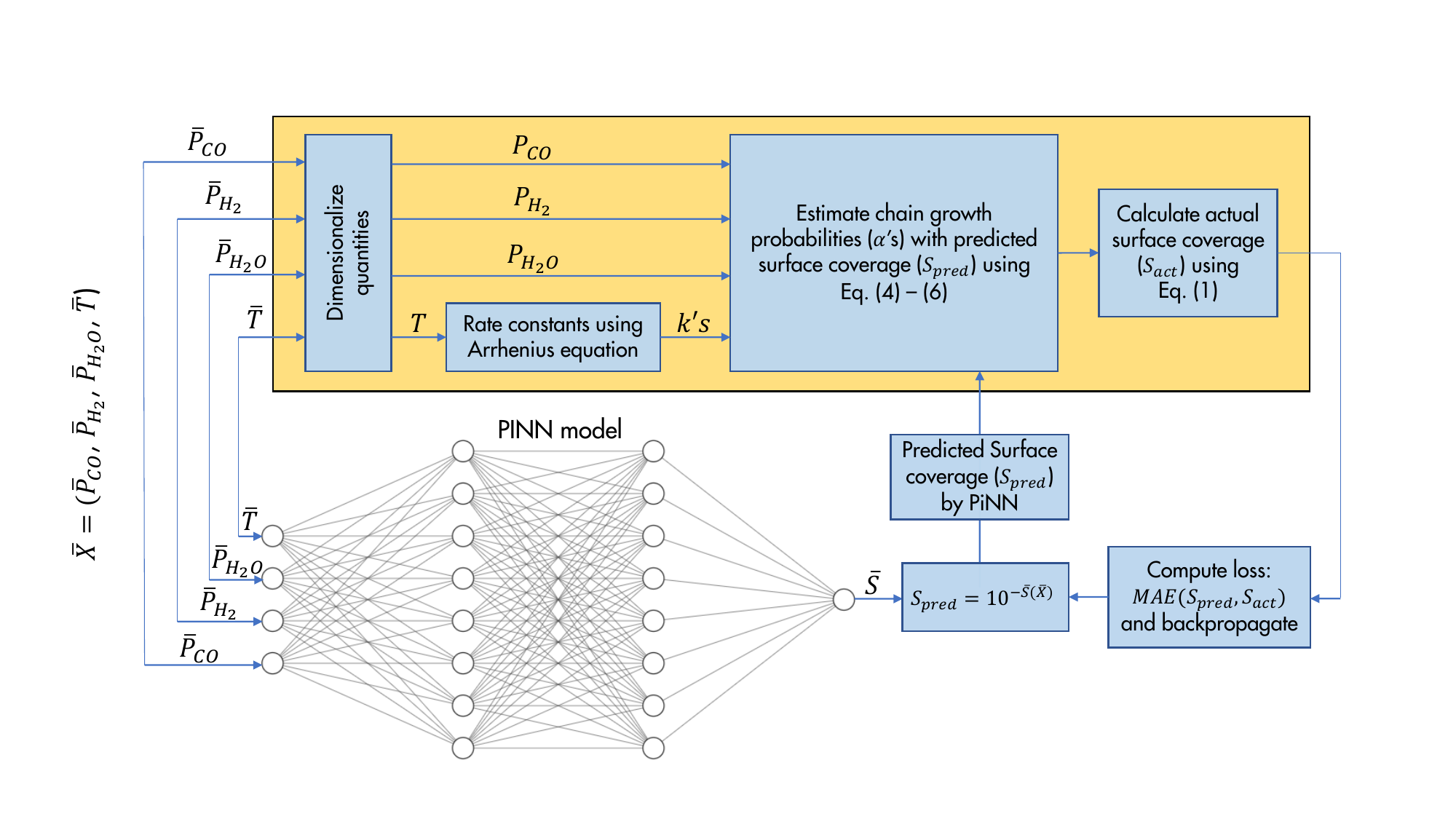}
    \caption{
    \label{fig:PINN_sch} 
    Block-scheme of PINN training process.
    }
\end{figure*}

The workflow of physics-informed NN training used in the proposed model is outlined in Fig. \ref{fig:PINN_sch}. Instead of relying on a number of input-output pairs, PINN training procedure starts with randomly generating a number of input instances $\bar{X}_j$ (its components are denoted as $\bar{P}_{CO}$, $\bar{P}_{H_2}$, $\bar{P}_{H_2 O}$ and $\bar{T}$ in Fig. \ref{fig:PINN_sch}) from the parametric ranges. These values are then simultaneously used as inputs into a) the NN being trained and b) are converted back into dimensional form and substituted into the target equations \ref{eq:S-equation}--\ref{eq:alpha_3n}. Dependence of temperature comes into effect from Arrhenius-type equations used to calculate reaction rate constants in equations \ref{eq:c_0}--\ref{eq:alpha_3n}.

In order to make the following discussion more precise, it is convenient to distinguish two types of $[S]$ quantity. The first one is given by equation \ref{eq:S_eq_10_Sbar} with $\bar{S}$ obtained as a prediction from the NN being trained. In Fig. \ref{fig:PINN_sch} we denote this quantity as $ S_{pred} = 10 ^ {-\bar{S}(\bar{X})} $. The second one i.e. $S_{act}$ is obtained when $ S_{pred}$ is substituted into equations \ref{eq:alpha_1}--\ref{eq:alpha_3n} and the resulting chain growth probabilities are used to evaluate r.h.s. of equation \ref{eq:S-equation}.

If PINN would have already been trained, $ S_{pred}$ given by it would satisfy equation \ref{eq:S-equation} exactly, and thus $ S_{pred}$ and $S_{act}$  would be equal. We thus use the discrepancy between these quantities to guide the training process. To this end, we define the loss function as the mean absolute error(MAE) between $S_{pred}$ and $S_{act}$ defined over all input instances $\bar{X}$. This loss function is minimized during the training process and in this way the set of optimal weights and biases is obtained.

\subsection{Implementation details}

The physics-informed NN training process outlined in the previous subsection was implemented by minimizing the Mean Absolute Error(MAE) loss function $E (\{ W^i, b^i \}) $ defined as
$$
E = \frac{1}{N_B} \sum\limits_{k = 1}^{N_B} \left|
10^{-\bar{S}(\bar{X}_k)}
-
\frac{1}{
c_0 + c_{S} \cdot
  \left(
     \sum\limits_{i=1}^{N_{max}} \prod\limits_{j=1}^i \alpha_j + \delta
   \right)
  }
\right| \; ,
$$
where $k$ enumerates the sampled points, 
$N_B$ is the batch size, $c_0$ and $c_{S}$ are given by equation \ref{eq:c_0} and \ref{eq:c_S}, $\delta$ is the `to-infinity' summation correction described in \ref{sec:to_inf_correction}, $N_{max} = 100$, $\alpha_j$ depends on $\bar{X}_k$ through equations \ref{eq:alpha_1}--\ref{eq:alpha_3n}, \ref{eq:X_to_Xbar}--\ref{eq:NN_as_eq}, and $ W^i$, $b^i$ were not indicated explicitly as the arguments of $E$ and $\bar{S}$ (given by equation \ref{eq:NN_as_eq}) for the sake of brevity.

All computations related to PINN were performed with a Python program developed as a part of the present study, using DeepXDE (version 1.3.1) library for scientific machine learning and physics-informed learning \cite{Lu-deepxde-2021} with TensorFlow \cite{Abadi-tensorflow-2016} backend.
\begin{table}
    \caption{\label{tbl:selected_hyperparameters} Hyperparameters selected
    for PINN model training}
    \centering{}%
    \begin{tabular}{|c|c|}
    \hline
    Hyperparameter & Value \\ \hline
    Training samples & 10000 \\ \hline
    Batch size & 10000 \\ \hline
    Epochs & 1000000 \\ \hline
    Resampling strategy & Random \\ \hline
    Resampling frequency & After each epoch \\ \hline
    Activation function & ReLU \\ \hline
    Weights initialization & Glorot uniform \\ \hline
    Initial learning rate & 0.001 \\ \hline
    Learning rate decay & Inverse time \\ \hline
    Decay steps & 100 \\ \hline
    Decay rate & 0.01 \\ \hline
    Optimizer & Adam \\ \hline
    Loss & Mean Absolute Error (MAE) \\ \hline
    Precision & Double \\ \hline
    \end{tabular}
\end{table}
The values of hyperparameters used to set-up the PINN and its training process are presented in Table \ref{tbl:selected_hyperparameters}, whereas the procedure used to select them is described in \ref{sec:hyperparameter_tuning}. The input points $\bar{X}_k$ used for the PINN training were sampled randomly from a $[0, 1]^4$ hypercube using DeepXDE's pseudo random type of sampling sequence. The probability distribution used to sample the points was adjusted dynamically after each epoch as the training proceeded, using an instance of  DeepXDE's \texttt{dde.callbacks.PDEResidualResampler} class. It automatically samples more points around those $\bar{X}_k$'s for which error/loss is more. This concept is similar to adaptive mesh refinement used for traditional mesh-based numerical solvers. Note that in newer DeepXDE versions, e.g., 1.7.2, the similar training results can be achieved by using \texttt{dde.callbacks.PDEPointResampler} class with \texttt{pde\_points} parameter set to \texttt{True}. The model showing the lowest value of the loss function $E$ during the training process was saved and used for further benchmarking of PINN performance. The same PINN solver was also implemented using NVIDIA Modulus (formerly known as NVIDIA SimNet \cite{Hennigh-NVIDIA-2021}) leading to similar results. Due to this, we report only the results obtained from DeepXDE-based implementation.

PINN training was performed on a 6-core Intel Xeon E5-2690 v4 CPU (2.60 GHz) workstation equipped with 110Gb RAM and NVIDIA Tesla V100-PCIE-16GB GPU. The same hardware configuration was also used to run some of the tests evaluating the speed of PINN computations at the inference stage. To specify this, we'll refer this system as the `workstation hardware' for brevity. Additional benchmarkings were performed on a portable notebook with 6-core Intel Core i7-8750H CPU, 32 GB DDR4 RAM and NVIDIA Quadro P2000 GPU. This configuration will be referred as the `notebook hardware'.

Unless otherwise specified, a standard \texttt{fsolve} routine from SciPy library \cite{Virtanen-SciPy-2020} (version 1.10.0) was used as the conventional solver for obtaining the `ground truth' values of $[S]$ when evaluating the accuracy and speedup of the proposed PINN model. In later experiments, MATLAB \cite{Higham-Matlab-2016} version 2021a was also used for benchmarking.

\section{Results and discussion}

Performance of the trained PINN model has been evaluated not only in terms of accuracy of the chemically relevant quantities (e.g., fraction of vacant sites, chain growth probabilities and reaction rates), but also from the perspective of its computational effectiveness (speedup) as compared to the conventional solver. We thus report these results separately in the subsections \ref{sec:accuracy-metrics}  and \ref{sec:time-metrics}.

Two sets of numerical experiments were performed for each of these subsections: (a) using the parameters of the kinetic model reported in Table 2 of \cite{Todic-CO-insertion-2014} and second, (b) using the parameters of a proprietary catalyst developed by Shell plc. Because the PINN accuracy metrics for the Shell catalyst were found to be worse, we report them in subsection \ref{sec:accuracy-metrics} when discussing the model accuracy. In terms of computational effectiveness (speedup), the model performance was found to be similar for both sets of experiments. Hence, for consistency, the speed-ups reported in subsection \ref{sec:time-metrics} also refers to the case of Shell catalyst.

In all the cases throughout this section, the benchmarking exercise was performed with the PINN model based on the fully-connected NN having 2 hidden layers with 512 neurons each (the `512~x~2' architecture). Details of different hyperparameters selection is present in \ref{sec:hyperparameter_tuning} whereas \ref{sec:more_speedups} covers accuracy and speedup benchmarks for different architectures and compute hardwares.

\subsection{Accuracy metrics}
\label{sec:accuracy-metrics}

To evaluate the accuracy of the proposed PINN model, we created a test dataset containing $18^4 = 104~976$ unique values of $\bar{X}$. It corresponds to all possible combinations of 18 values for each of its four components, placed uniformly on a $(0, 1)$ range as $\frac{k}{19}$, where $k = 1, 2, ..., 18$. For each of these inputs, the `ground truth' values of $[S]$ and  $\alpha_n$ were computed using a conventional solver. All values of $[S]$ and $\alpha_n$, obtained either from the PINN model or from the conventional solver, were further converted into the reaction rates for paraffins ($R_{\mathrm{C}_n \mathrm{H}_{2 n + 2}}$) and 1-olefins ($R_{\mathrm{C}_n \mathrm{H}_{2 n}}$) using equations \ref{eq:R_CH4}--\ref{eq:R_C2H2n}.

The difference between the predictions obtained using PINNs and the ground truth values for all of the above quantities was evaluated by a relative error metric
$$
\epsilon_{x} = \frac{|x_{PINN} - x_{true}|}{|x_{true}|} \cdot 100 \% \; ,
$$
where $x$ stands for $[S]$, $\alpha_n$, $R_{\mathrm{C}_n \mathrm{H}_{2 n + 2}}$ or $R_{\mathrm{C}_n \mathrm{H}_{2 n}}$. In case of the reaction rates, relative errors are for $n=N_{max}=100$ as error increases with increase in carbon number due to the nature of reaction rates equations \ref{eq:R_CH4}--\ref{eq:R_C2H2n}. $\epsilon_{x}$ was obtained for each of the 104~976 test points and from these errors, three statistical descriptors were finally calculated: mean, median and maximum relative error. Obtained metrics are presented in Table \ref{tbl:accuracy_metrics_summary}.  Similarly, values for several additional NN architectures are reported in \ref{sec:hyperparameter_tuning} (Table \ref{tbl:model_accuracy}).

\begin{table}
    \caption{\label{tbl:accuracy_metrics_summary} Relative error metrics of the chemically relevant quantities produced by the proposed PINN model with 512~x~2 architecture
    }
    \begin{tabular}{|c|c|c|c|}
      \hline
      \multirow{2}{*}{Quantity} & \multicolumn{3}{c|}{ Relative error (in \%)} 
      \\
      \cline{2-4}
      & Mean & Median & Max\\
      \hline
      $[S]$ & 4.60E-02 & 3.35E-02 & 8.56E-01\\
      \hline
      $\alpha_n$ & 3.35E-03 & 1.09E-03 & 3.14E-01\\
      \hline
      $R_{\mathrm{C}_n \mathrm{H}_{2 n + 2}}$ & 1.13E-01 & 5.67E-02 & 7.64E+00\\
      \hline
      $R_{\mathrm{C}_n \mathrm{H}_{2 n}}$ & 1.59E-01 & 9.13E-02 & 8.35E+00\\
      \hline
    \end{tabular}
\end{table}

\begin{figure}
    \begin{centering}
        \includegraphics[width=\linewidth]{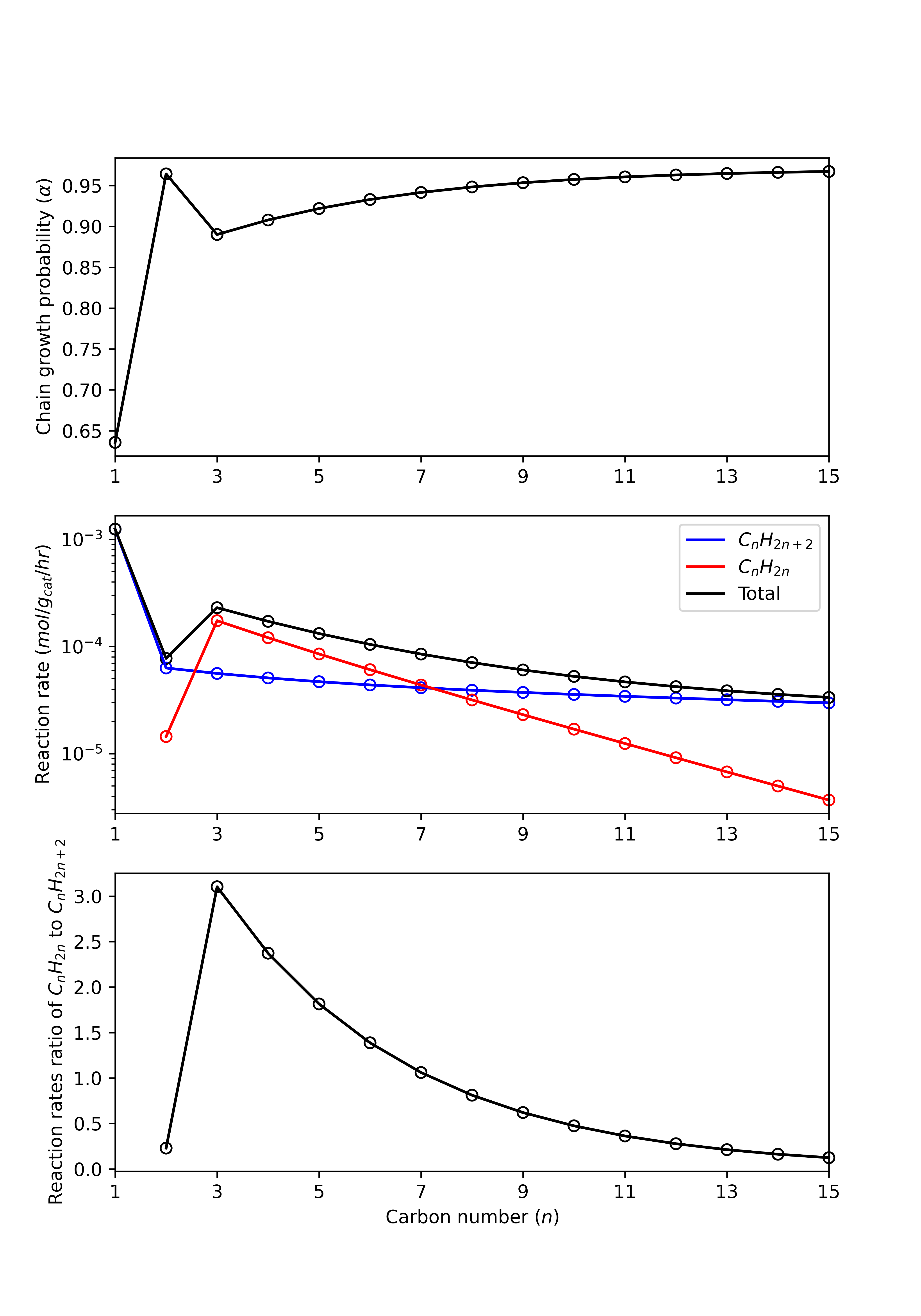}
        \par\end{centering}
    \caption{\label{fig:alphas_rr_plot} Plots of top: chain growth probability ($\alpha$), middle: reaction rates and bottom: reaction rate ratio of $C_{n}H_{2n}$ to $C_{n}H_{2n+2}$ with carbon chain length ($n$) using PiNN (-) and root finding ($\circ$) at $P_{CO}=1$ MPa, $P_{H_{2}}=1$ MPa, $P_{H_{2}O}=0.5$ MPa and $T=493.15$ K.}
\end{figure}

Fig. \ref{fig:alphas_rr_plot} presents the visual comparison between the results obtained from the proposed PINN model and the ground truth values for a) chain growth probabilities, b) absolute values of paraffin and 1-olefins reaction rates and c) olefin-to-paraffin ratio $\frac{ R_{\mathrm{C}_n \mathrm{H}_{2 n}} }{ R_{\mathrm{C}_n \mathrm{H}_{2 n + 2}} }$ for different carbon chain lengths. For this experiment, we used kinetic model parameters reported in Table 2 of \cite{Todic-CO-insertion-2014}. Presented plots confirm good agreement between the results produced by our PINN model and their ground truth values. It can thus be concluded that the accuracy of all chemically relevant quantities produced by the proposed PINN model should be sufficient for their practical applications.

\subsection{Computational effectiveness}
\label{sec:time-metrics}

One of the benefits of a PINN model over a conventional solver is the ease of porting PINN-based computations to GPU. Potential speed-up achieved by leveraging GPU comes from the type of computations NN relies on. These are predominantly matrix-vector manipulations, in which the same elementary operations have to be executed with many different variables and which are thus very well suited for parallelization (basically, by the so-called single instruction, multiple data (SIMD) type of parallel computing). Another characteristic of computations used by NN workflow is a very small number of branching steps or even the absence of such. This is in contrast to conventional solvers which usually rely on iterative root finding algorithms (such as bracketing, Newton-Raphson, secant, bisection methods) in which the flow of the program can significantly vary at runtime. We thus expect high speed-ups when comparing PINN model running on GPU as compared to a traditional solver.

To verify these considerations, we measured wall-time $t_{PINN}(N_{batch})$ required by the proposed PINN model to compute $[S]$ for the test dataset with different batch sizes ($N_{batch}$). In these experiments, the test data points ($\bar{X}$) were sampled randomly from uniform distribution on $[0, 1]$ interval. To get reliable estimates of $t_{PINN}(N_{batch})$, it was averaged over at least 1000 inference runs. Time required by the conventional solver was calculated as $N_{batch} \cdot t_{conv}$ to compute the same number of $[S]$ values on the test dataset. Using these compute times, the PINN speed-up factor can be calculated as the ratio $\frac{N_{batch} \cdot t_{conv}}{t_{PINN}(N_{batch})}$.

Two sets of numerical experiments were performed. In the first set of experiments, Python SciPy \texttt{fsolve} with default settings was used as a conventional solver, having the average $t_{conv} = 0.1014$ sec while running on a single-thread of the notebook CPU. Obtained speed-up factors are presented in Table \ref{tbl:Speedups_compared_to_scipy_fsolve} for 512~x~2 PINN architecture, while similar results for a number of other PINN architecture are presented in \ref{sec:more_speedups}. Results imply that the observed speed-up depends on the number of samples which the PINN model processes simultaneously (the batch sizes). Initially, the speed-up increases with the batch size and slowly plateaus for the larger batches as the GPU gets fully utilized. On both GPUs used for benchmarking, the speed-up factor hits a plateau after the batch size reaches ${\sim} 10^4$. For small batch sizes, the overheads related to data transfer between CPU and GPU memory and CPU-GPU synchronizations become significant and downgrades the overall performance, while for large batches such delays become insignificant compared to the compute time. However, even for small batch sizes, ${\sim} 100$ times speed-up is readily achievable.

\begin{table}
    \caption{\label{tbl:Speedups_compared_to_scipy_fsolve} 
    Speed-up factor achieved using the proposed PINN model on the workstation GPU (NVIDIA Tesla V100) and the notebook GPU (NVIDIA Quadro P2000) with respect to the Python SciPy \texttt{fsolve} based conventional solver}

    \begin{tabular}{|c|c|c|}
      \hline
      Batch size & Workstation & Notebook \\
      \hline
      $10^{0}$ & 1.81E+02 & 8.76E+01\\
      \hline
      $10^{1}$ & 1.82E+03 & 8.35E+02\\
      \hline
      $10^{2}$ & 1.85E+04 & 3.38E+03\\
      \hline
      $10^{3}$ & 1.61E+05 & 9.64E+03\\
      \hline
      $10^{4}$ & 5.12E+05 & 1.22E+04\\
      \hline
      $10^{5}$ & 7.02E+05 & 1.24E+04\\
      \hline
      $10^{6}$ & 7.37E+05 & (Out-of-memory)\\
      \hline
    \end{tabular}
\end{table}

In the second set of experiments, the goal was to evaluate the performance of the trained PINN model inside a MATLAB environment. For this task, we used the notebook hardware with relatively less powerful NVIDIA Qurdo P2000 GPU. We expect the proposed PINN model will be used as a component of a more advanced FTS optimization workflows, which is likely to be developed and used by chemical engineers in a more familiar MATLAB environment, running on a portable computer.

Although both MATLAB and Python are interpreted languages, they differ in their design and implementation approaches in delegating time-critical computations to an external code, which is typically compiled from C, C++ or Fortran sources. MATLAB, being a commercial offering, generally has more efficient underlying subroutines to perform compute intensive tasks. So not so surprisingly, proprietary MATLAB \texttt{fsolve} with default settings took just $t_{conv} = 1.53\cdot 10^{-3}$ sec on average to compute value of $[S]$ by solving equations \ref{eq:S-equation}--\ref{eq:alpha_3n}. This is around 66 times faster than its Python SciPy \texttt{fsolve} counterpart, and can be attributed to the differences in MATLAB solver implementation, which uses C/C++ compiled executable files with vectorization techniques to speedup its proprietary \texttt{fsolve} function.

MATLAB environment may also offer further speed-ups in the PINN model inference due to its compute efficiency. To this end, the trained PINN in TensorFlow was converted to ONNX format (the open standard for machine learning models interoperability) \cite{Sawarkar-deep-2022, Shridhar-interoperating-2020}
and then imported into a MATLAB program to perform inference and evaluate its execution time. The inference was done in two different ways: 1) via a standard MATLAB \texttt{predict} function and 2) by using MATLAB's GPU Coder to generate optimized C++ code capable of taking advantage of cuDNN \cite{Chetlur-cudnn-2014} (a GPU-accelerated library of primitives for deep neural networks running on NVIDIA GPUs supporting CUDA). In the second case, the generated code was compiled into a MATLAB executable (MEX file), which was then called from inside the MATLAB environment.

Speed-up factors of MATLAB PINN inference as compared to the conventional solver based on its proprietary \texttt{fsolve} function are presented in Table \ref{tbl:model_performance_matlab-ref_P2000}. It can be seen from these results that the speed-up growths as the batch size increases, similarly to the experiments of PINN benchmarking in Python. Additional speed-up is achieved using the MEX compilation, which makes PINN inference nearly an order of magnitude faster. Overall, the speed-up in MATLAB environment can be as high as ${\sim} 10^3$ (in MEX mode), saturating at batch size of ${\sim} 10^3$. Note that this speed up is achieved on a low end NVIDIA P2000 notebook GPU. With workstation category GPUs, the speed up is expected to be 10-100 time better.

\begin{table}
    \caption{\label{tbl:model_performance_matlab-ref_P2000} Speed-up factors of MATLAB PINN inference (with or without MEX compilation) on notebook hardware with respect to its propritory \texttt{fsolve} based conventional solver 
    }
    \begin{tabular}{|c|c|c|}
      \hline
      Batch size & Without MEX & With MEX\\
      \hline
      $10^0$ & 6.81E-01 & 9.01E+00\\
      \hline
      $10^1$ & 4.17E+00 & 9.35E+01\\
      \hline
      $10^2$ & 4.72E+01 & 6.72E+02\\
      \hline
      $10^3$ & 1.45E+02 & 1.42E+03\\
      \hline
      $10^4$ & 1.55E+02 & 1.53E+03\\
      \hline
      $10^5$ & 1.64E+02 & 1.67E+03\\
      \hline
      $10^6$ & 7.68E+01 & 1.65E+03\\
      \hline
    \end{tabular}
\end{table}

Interestingly, it has been noticed that the inference time of the PINN model is reduced when MATLAB is used as its execution environment instead of a Python. In order to evaluate this, we measured additional speed-up factors, defined as the ratio ${t_{PINN}^{Python} } / {t_{PINN}^{MATLAB}}$. Here, $t_{PINN}^{Python}$ and $t_{PINN}^{MATLAB}$ are the PINN model inference times when running it on Python and MATLAB environments, respectively. Note that MATLAB environment is evaluated with and without MEX compilation mode. In both cases notebook GPU (NVIDIA Qurdo P2000) was used for inference. Resulting speed-up factors are presented in Table \ref{tbl:model_performance_py-ref_P2000}. It can be seen from the obtained results that when MEX compilation is not used, PINN inference in MATLAB environment performs a bit slower than Python. However, MEX compilation allows $\sim$ 6 to 13 times faster inference in MATLAB environment, with $10^2$ being an optimal batch size, although the dependency on batch size is not trivial.

\begin{table}
    \caption{\label{tbl:model_performance_py-ref_P2000} Speed-up factors ${t_{PINN}^{Python} } / {t_{PINN}^{MATLAB}}$
    of PINN inference (with or without MEX compilation in MATLAB) on notebook hardware
    }
        \begin{tabular}{|c|c|c|}
      \hline
      Batch size & Without MEX & With MEX\\
      \hline
      $10^0$ & 5.14E-01 & 6.80E+00\\
      \hline
      $10^1$ & 3.30E-01 & 7.41E+00\\
      \hline
      $10^2$ & 9.24E-01 & 1.32E+01\\
      \hline
      $10^3$ & 9.96E-01 & 9.72E+00\\
      \hline
      $10^4$ & 8.45E-01 & 8.29E+00\\
      \hline
      $10^5$ & 8.76E-01 & 8.93E+00\\
      \hline
    \end{tabular}
    \end{table}

\section{Conclusions and Outlook}

We demonstrated that a physics-informed neural network (PINN) can be used to solve highly coupled non-linear equations describing a wide class of microkinetics models of Fischer-Tropsch synthesis process.
In the current work, we show that a two-layer fully-connected feed-forward network achieves relative median error of 0.03\% in predicting the fraction of vacant catalytic sites as a function of the reaction conditions, given by partial pressures of $\mathrm{H}_2$, $\mathrm{CO}$ and $\mathrm{H}_2 \mathrm{O}$ and temperature. One of the major benefits of using the PINN method is that it can achieve up to $7\cdot10^5$ speed-up when running the trained model for inferencing on GPU, as compared to a conventional equation solver. PINN method thus achieves an excellent speed up while maintaining accuracy comparable with the conventional solvers. Such an ultra-fast forward model can be utilised for many downstream digital applications. This model can be used as it is, for digital twin application where such a forward simulator is needed as a part of a bigger workflow that predicts the product formation rates or the performance of the catalyst in terms of selectivity and yield. Another application of such a fast and accurate model will be within an optimization loop where this model acts as a forward simulator while complimenting a separate optimizer whose task is to find the optimal process conditions to match a user specified product slate. This integration with a gradient based optimizer can be made even more tighter while benefiting from additional speedup coming from the gradients that are available from the neural network almost free of (computational) cost. It comes from the fact that a PINN model converts the implicitly coupled system of equations into explicitly connected NN layers which allows for the automatic differentiation over the trained PINN model, thereby providing the derivatives of outputs (such as chain growth probabilities, reaction rates etc) with respect to the inputs (process conditions).

While in this work the thermodynamic and the reaction kinetics parameters are set as known constants, these can also be added as the inputs to the model, thus making them additional parameters. A situation where this approach might be needed is when we have the reaction rates or selectivity and yield data obtained from a lab scale micro-reactor by varying the process conditions. In such a case PINN inverse problem could be framed with a gradient based optimizer to find the set of rection kinetics and thermodynamic parameters that best matches the catalyst performance observed in such experiments. This is one of the kind of inverse problems where a mathematical prior (expressed in terms of equations representing the microkinetic model) is used to fine-tune the thermodynamic and kinetic parameters for the specific catalyst of interest.

PINN-based microkinetics models can also be very important for the multi-scale and multi-physics reactor modelling. They can be used in the pellet scale or reactor scale models where the different transport phenomena (eg. fluid flow, mass transfer, heat transfer) are also taken into account. Reaction rates of the constituent species from such kind of model appear as source terms in the convection-diffusion equations for the species and temperature field. The predictions obtained using our PINN-based approach for microkinetics can thus be used along with a conventional solver or can be glued together with another set of PINNs in a hierarchical PINNs based approach. Such a hierarchical approach would comprise of one set of PINNs learning the diffusion equations within the pellets while querying the PINN model presented in this work to fetch the reaction rates that are modelled as source terms. The resulting system would allow to quantify the pellet performance at different reaction conditions and observe any mass transfer limitations occurring in the system.

\section*{Acknowledgements}
\label{sec:ack}
The authors would like to thank Shell plc, specifically computational science and digital innovation leadership team for funding this R\&D work. We also sincerely acknowledge and appreciate the guidance and direction offered by Dan Jeavons, Detlef Hohl, Suchismita Sanyal, Vibhor Aggarwal, Amy Challen and Karina Fernandez which resulted into this publication.


\appendix

\section{`To-infinity' summation correction}
\label{sec:to_inf_correction}

The procedure used to derive equation \ref{eq:S-equation} imposes no limitation on the maximum value of carbon chain-length ($n$) in the products $\prod_{j=1}^n \alpha_j$ being summed in the ultimate balance equation. Ideally, we'd like to consider all possible values of $n$, i.e., sum-up such products up to infinity (note that $\alpha_n \le 1$, so that these products will become small allowing for a finite result of the summation). However, to make such infinite summation computationally tractable, certain approximations are needed.

Consider $n \ge N_0$ and suppose that starting from this $N_0$, the chain growth probability becomes independent of the carbon chain-length ($n$), i.e., $\alpha_j \approx \alpha_{N_0}$ for all $j \ge N_0$. 
This happens because after some $n$ the $e^{c\cdot n}$ multiplier in equation \ref{eq:alpha_3n} will become negligibly small.
Then for $n \ge N_0$
\begin{eqnarray*}
  \prod_{j = 1}^n \alpha_j & = & \prod_{j = 1}^{N_0} \alpha_j \cdot \prod_{j =
  N_0 + 1}^n \alpha_j\\
  & \approx & \prod_{j = 1}^{N_0} \alpha_j \cdot (\alpha_{N_0})^{n -
  N_0}\\
  & = & \frac{\prod_{j = 1}^{N_0} \alpha_j}{(\alpha_{N_0})^{N_0}} \cdot
  (\alpha_{N_0})^n
\end{eqnarray*}
Given that, we can compute a `to-infinity' summation correction
$$
\delta = \sum_{n = 1}^{\infty} \left( \prod_{j = 1}^n \alpha_j \right) -
   \sum_{n = 1}^{N_0} \left( \prod_{j = 1}^n \alpha_j \right) = \sum_{n = N_0
   + 1}^{\infty} \left( \prod_{j = 1}^n \alpha_j \right)
$$   
It is defined as the sum of the terms which are `missing' in the finite summation
$\sum_{n = 1}^{N_0} \left( \prod_{j = 1}^n \alpha_j \right)$
when compared to the infinite sum
$\sum_{n = 1}^{\infty} \left( \prod_{j = 1}^n \alpha_j \right)$.
We thus find
\begin{eqnarray*}
  \sum_{n = N_0 + 1}^{\infty} \left( \prod_{j = 1}^n \alpha_j \right) &
  \approx & \frac{\prod_{j = 1}^{N_0} \alpha_j}{(\alpha_{N_0})^{N_0}}
  \cdot \sum_{n = N_0 + 1}^{\infty} (\alpha_{N_0})^n\\
  & = & \frac{\prod_{j = 1}^{N_0} \alpha_j}{(\alpha_{N_0})^{N_0}} \cdot
  \sum_{k = 1}^{\infty} (\alpha_{N_0})^{N_0 + k}\\
  & = & \prod_{j = 1}^{N_0} \alpha_j \cdot \; \sum_{k =
  1}^{\infty} (\alpha_{N_0})^k
\end{eqnarray*}
Here $\sum_{k = 1}^{\infty} (\alpha_{N_0})^k = \frac{\alpha_{N_0}}{1 -
\alpha_{N_0}}$ as the sum of geometric series. Note that $\alpha_j \le
1$ by their definition, which guarantees the convergence of the series.

Finally,
$$
\delta \approx \prod_{j = 1}^{N_0} \alpha_j \cdot \frac{\alpha_{N_0}}{1 - \alpha_{N_0}} \; ,
$$
where we put $N_0 = N_{max} = 100$.

\section{Hyperparameter Tuning}
\label{sec:hyperparameter_tuning}

Accuracy of the proposed approach relies mostly on the performance of neural network(NN), which is its core ingredient. That's why substantial attention was paid to optimizing the architecture of this network and training procedure. Collectively, these settings of the learning process are usually referred as hyperparameters. The number of all possible combinations of individual hyperparameters values is generally too large to be enumerated in a brute force manner. Therefore, we organized the process of (manual) hyperparameters tuning in the following way.

After initial explorations, we selected MAE as a loss function and took 512~x~2 architecture (that is, 512 neurons in each of the two hidden layers of a fully connected NN) as an initial guess to start the process of hyperparameters tuning. We evaluated the effect of the batch size, optimization algorithm (`optimizer') selection, initial learning rate and learning rate decay. The latter controlled the learning rate scheduler which we selected to be the so-called `inverse time decay'. It prescribes the following dependence of learning rate $\eta$ on the number of epoch $t$
$$
\eta = \frac{\eta_0} {1 + \kappa \cdot \frac{t}{t_s} } \; ,
$$
where $\kappa$ is the learning rate decay, $\eta_0$ is the initial learning rate and $t_s = 100$.
The total number of training epochs was kept fixed at 1~000~000 in all experiments, and the weights corresponding to the lowest value of loss function observed during entire training process on training data points was used as the final result. The `ground truth' values of $[S]$ were thus never used for determining the epoch with the `best' set of weights. However, such values were used when evaluating the accuracy of the trained models and guide the selection of the best-performing hyperparameters. To this end, test batch of $18^4 = 104~976$ uniformly sampled points from the parameter ranges was generated and the value of $[S]$ was predicted using PINN model for each of these samples. Relative (\%) errors in the PINN predictions were computed with respect to the numerical roots calculated using Python SciPy \texttt{fsolve}.

When effect of one of the hyperparameters was evaluated, the remaining ones were set to: $\eta_0 = 10^{-3}$ (initial learning rate), $\kappa = 0.01$ (learning rate decay), 10~000 (batch size), Adam \cite{Kingma-adam-2017}  (optimizer). Starting from this combination, we assessed three values of the initial learning rates ($10^{-2}$, $10^{-3}$, $10^{-4}$, see Table \ref{tbl:initial_lr}), three values of the learning rate decay ($10^{-1}$, $10^{-2}$, $10^{-3}$, see Table \ref{tbl:lr_decay}), four values of the batch size used for training (1000, 3000, 10000, 30000, see Table \ref{tbl:batch_size}) and four options for the optimization algorithm (see Table \ref{tbl:optimizer_selection}). Three of them were Adam, RMSProp and L-BFGS (see \cite{Aggarwal-training-2018, Tian-recent-2023, He-large-scale-2021, Guo-overview-2023} for recent reviews), and one more option was to use Adam optimizer during the first 200~000 epochs and then switching to L-BFGS for rest of the epochs (`Adam + L-BFGS' in Table \ref{tbl:optimizer_selection}).
Default parameters of all optimizers were used as implemented in TensorFlow. As a result, it has been confirmed that the combination of initial learning rate, learning rate decay, batch size and optimizer shown in Table \ref{tbl:selected_hyperparameters} leads to the best accuracy, in a sense that the change in any of these parameters alone within the tested ranges does not improve the accuracy.

After finding the best hyperparameters of the NN training process, we experimented with the NN architecture. Because we fixed the type of NN i.e. fully connected feed-forward NN, the only hyperparameters of its architecture were the number of neurons in the hidden layers and the number of hidden layers. We evaluated the architecture with 32, 64, 128, 256, 512 neurons in the hidden layers, and used 2, 3 or 4 layers in the network. Results are given in Table \ref{tbl:model_accuracy}. It can be seen that the `512~x~2' architecture yields the best overall results.

The final set of experiments were performed to evaluate the effect of NN training points sampling strategy on the model accuracy. This was done using DeepXDE 1.7.2 by using two possible options for \texttt{pde\_points} parameter of \texttt{dde.callbacks.PDEPointResampler} class. When this parameter is set to \texttt{True} (this corresponds to the mode used to obtain all results reported in the main section of the paper), probability density for sampling random training points is proportional to the NN loss. So, higher number of points are sampled from those regions where the NN loss is more \cite{Lu-deepxde-2021}. Note that the training points are resampled after each epoch using the loss function. Results presented in Table \ref{tbl:floating_point_precision} indicate that such resampling is quite important, because it drastically reduces the median relative error in $[S]$ almost by the factor of 5.

The same experiment was also performed when the default double-precision representation of floating point numbers (used to produce all results in the main part of the present paper) was changed to a single-precision one. It can be seen that such change did not downgrade the model accuracy noticeably, but allowed training to be almost two times faster.
Moreover, switching to single precision representation allows further speed-up of model inference by a factor of  ${\sim} 2$. This option has not been used during speed-ups benchmarkings however, because otherwise, conventional solvers should be run in single precision mode as well, which may not be their native mode of operation.

\begin{table*}
    \caption{\label{tbl:initial_lr} Effect of initial learning rate}
    \centering{}%
    \begin{tabular}{|c|c|c|c|c|c|}
    \hline
    \multirow{2}{*}{Initial learning rate} & \multirow{2}{*}{Train loss} & \multirow{2}{*}{Test loss} & \multicolumn{3}{c|}{Relative (\%) error in S} \\ 
    \cline{4-6} & & & Mean & Median & Max \\ \hline
    0.01 & 6.55E-05 & 4.83E-05 & 5.17E-01 & 4.89E-01 & 2.22E+00 \\ \hline
    0.001 & 3.58E-05 & 2.82E-05 & 4.60E-02 & 3.35E-02 & 8.56E-01 \\ \hline
    0.0001 & 1.11E-04 & 7.33E-05 & 5.30E-01 & 4.91E-01 & 2.90E+00 \\ \hline
    \end{tabular}
\end{table*}

\begin{table*}
    \caption{\label{tbl:lr_decay} Effect of learning rate decay}
	\centering{}%
    \begin{tabular}{|c|c|c|c|c|c|}
    \hline
    \multirow{2}{*}{Learning rate decay} & \multirow{2}{*}{Train loss} & \multirow{2}{*}{Test loss} & \multicolumn{3}{c|}{Relative (\%) error in S} \\ 
    \cline{4-6} & & & Mean & Median & Max \\ \hline
    0.1 & 8.03E-05 & 5.89E-05 & 5.23E-01 & 4.90E-01 & 2.62E+00 \\ \hline
    0.01 & 3.58E-05 & 2.82E-05 & 4.60E-02 & 3.35E-02 & 8.56E-01 \\ \hline
    0.001 & 4.28E-05 & 3.54E-05 & 5.12E-01 & 4.88E-01 & 1.57E+00 \\ \hline
    \end{tabular}
\end{table*}

\begin{table*}
    \caption{\label{tbl:batch_size} Effect of batch size}
	\centering{}%
    \begin{tabular}{|c|c|c|c|c|c|}
    \hline
    \multirow{2}{*}{Batch size} & \multirow{2}{*}{Train loss} & \multirow{2}{*}{Test loss} & \multicolumn{3}{c|}{Relative (\%) error in S} \\ 
    \cline{4-6} & & & Mean & Median & Max \\ \hline
    1000 & 4.45E-05 & 3.98E-05 & 6.69E-02 & 4.90E-02 & 8.90E-01 \\ \hline
    3000 & 4.50E-05 & 3.42E-05 & 5.48E-02 & 4.10E-02 & 8.28E-01 \\ \hline
    10000 & 3.58E-05 & 2.82E-05 & 4.60E-02 & 3.35E-02 & 8.56E-01 \\ \hline
    30000 & 3.51E-05 & 2.84E-05 & 4.68E-02 & 3.42E-02 & 8.12E-01 \\ \hline
    \end{tabular}
\end{table*}

\begin{table*}
    \caption{\label{tbl:optimizer_selection} Effect optimizer selection}
	\centering{}%
    \begin{tabular}{|c|c|c|c|c|c|}
    \hline
    \multirow{2}{*}{Optimizer} & \multirow{2}{*}{Train loss} & \multirow{2}{*}{Test loss} & \multicolumn{3}{c|}{Relative (\%) error in S} \\ 
    \cline{4-6} & & & Mean & Median & Max \\ \hline
    Adam & 3.58E-05 & 2.82E-05 & 4.60E-02 & 3.35E-02 & 8.56E-01 \\ \hline
    RMSProp & 7.91E-05 & 6.92E-05 & 5.24E-01 & 4.89E-01 & 2.06E+00 \\ \hline
    L-BFGS
    & 1.24E-03 & 9.03E-04 & 1.77E+00 & 1.10E+00 & 4.25E+01 \\ \hline
    Adam + L-BFGS
    & 3.34E-05 & 5.39E-05 & 5.18E-01 & 4.89E-01 & 2.16E+00 \\ \hline
    \end{tabular}
\end{table*}

\begin{table*}
    \caption{\label{tbl:floating_point_precision} Effect of floting point precision and PDE residual resampling}
	\centering{}%
    \begin{tabular}{|c|c|c|c|c|c|}
    \hline
    \multirow{2}{*}{Precision} & \multirow{2}{*}{PDE residual resampling} & \multirow{2}{*}{Train time (sec)} & \multicolumn{3}{c|}{Relative (\%) error in S} \\ 
    \cline{4-6} & & & Mean & Median & Max \\ \hline
    double & True & 22517 & 5.22E-02 & 3.83E-02 & 9.09E-01 \\ \hline
    double & False & 21916 & 3.16E-01 & 1.80E-01 & 4.93E+01 \\ \hline
    single & True & 12516 & 4.64E-02 & 3.48E-02 & 6.14E-01 \\ \hline
    single & False & 11898 & 3.03E-01 & 1.70E-01 & 1.95E+01 \\ \hline
    \end{tabular}
\end{table*}

\begin{sidewaystable}
\centering
\caption{\label{tbl:model_accuracy} Model Accuracy}
    \begin{adjustbox}{width=\columnwidth,center}
    \begin{tabular}{|c|c|c|c|c|c|c|c|c|c|c|c|c|c|c|}
    \hline
    \multirow{3}{*}{Architecture} & \multicolumn{2}{c|}{\multirow{2}{*}{Loss}} & \multicolumn{12}{c|}{Relative (\%) error} \\
    \cline{4-15} & \multicolumn{2}{c|}{} & \multicolumn{3}{c|}{$S$} & \multicolumn{3}{c|}{$\alpha$} & \multicolumn{3}{c|}{$RR_{C{{n}H_{2n+2}}}$} & \multicolumn{3}{c|}{$RR_{C{{n}H_{2n}}}$} \\
    \cline{2-15} & Train & Test & Mean & Median & Max & Mean & Median & Max & Mean & Median & Max & Mean & Median & Max \\ \hline
    [32]*2 & 7.75E-04 & 5.28E-04 & 1.04E+00 & 6.62E-01 & 1.83E+01 & 9.33E-02 & 2.10E-02 & 5.55E+00 & 3.18E+00 & 1.10E+00 & 2.81E+02 & 4.32E+00 & 1.78E+00 & 3.49E+02 \\ \hline
    [32]*3 & 4.01E-04 & 3.20E-04 & 5.92E-01 & 4.23E-01 & 8.44E+00 & 4.34E-02 & 1.38E-02 & 1.87E+00 & 1.46E+00 & 7.15E-01 & 5.60E+01 & 2.05E+00 & 1.15E+00 & 6.20E+01 \\ \hline
    [32]*4 & 3.00E-04 & 2.32E-04 & 4.39E-01 & 3.02E-01 & 1.04E+01 & 3.36E-02 & 9.77E-03 & 4.48E+00 & 1.12E+00 & 5.08E-01 & 2.09E+02 & 1.56E+00 & 8.18E-01 & 2.41E+02 \\ \hline
    [64]*2 & 3.67E-04 & 2.62E-04 & 4.76E-01 & 3.33E-01 & 5.30E+00 & 3.71E-02 & 1.05E-02 & 1.50E+00 & 1.22E+00 & 5.52E-01 & 3.87E+01 & 1.69E+00 & 8.94E-01 & 4.48E+01 \\ \hline
    [64]*3 & 2.17E-04 & 1.65E-04 & 3.19E-01 & 2.08E-01 & 6.09E+00 & 2.58E-02 & 6.64E-03 & 1.28E+00 & 8.41E-01 & 3.47E-01 & 3.27E+01 & 1.16E+00 & 5.61E-01 & 3.81E+01 \\ \hline
    [64]*4 & 1.79E-04 & 1.38E-04 & 2.41E-01 & 1.69E-01 & 5.10E+00 & 1.77E-02 & 5.48E-03 & 9.32E-01 & 5.98E-01 & 2.86E-01 & 2.36E+01 & 8.40E-01 & 4.60E-01 & 2.74E+01  \\ \hline
    [128]*2 & 1.53E-04 & 1.13E-04 & 2.12E-01 & 1.48E-01 & 3.21E+00 & 1.60E-02 & 4.79E-03 & 9.46E-01 & 5.37E-01 & 2.50E-01 & 2.67E+01 & 7.50E-01 & 4.03E-01 & 3.00E+01 \\ \hline
    [128]*3 & 1.05E-04 & 8.07E-05 & 1.44E-01 & 1.02E-01 & 2.99E+00 & 1.09E-02 & 3.35E-03 & 1.04E+00 & 3.63E-01 & 1.74E-01 & 2.36E+01 & 5.08E-01 & 2.80E-01 & 2.55E+01 \\ \hline
    [128]*4 & 1.01E-04 & 7.90E-05 & 1.45E-01 & 1.05E-01 & 3.31E+00 & 1.10E-02 & 3.41E-03 & 1.11E+00 & 3.67E-01 & 1.77E-01 & 3.13E+01 & 5.12E-01 & 2.86E-01 & 3.57E+01 \\ \hline
    [256]*2 & 7.29E-05 & 5.65E-05 & 1.05E-01 & 7.07E-02 & 1.99E+00 & 8.45E-03 & 2.25E-03 & 5.60E-01 & 2.75E-01 & 1.18E-01 & 1.47E+01 & 3.79E-01 & 1.91E-01 & 1.69E+01 \\ \hline
    [256]*3 & 6.39E-05 & 5.05E-05 & 8.66E-02 & 6.36E-02 & 1.19E+00 & 6.27E-03 & 2.05E-03 & 5.24E-01 & 2.13E-01 & 1.07E-01 & 1.27E+01 & 2.99E-01 & 1.73E-01 & 1.37E+01 \\ \hline
    [256]*4 & 5.57E-05 & 4.51E-05 & 7.65E-02 & 5.68E-02 & 1.26E+00 & 5.36E-03 & 1.83E-03 & 3.24E-01 & 1.84E-01 & 9.58E-02 & 8.57E+00 & 2.60E-01 & 1.55E-01 & 9.46E+00 \\ \hline
    [512]*2 & 3.58E-05 & 2.82E-05 & 4.60E-02 & 3.35E-02 & 8.56E-01 & 3.35E-03 & 1.09E-03 & 3.14E-01 & 1.13E-01 & 5.67E-02 & 7.64E+00 & 1.59E-01 & 9.13E-02 & 8.35E+00 \\ \hline
    [512]*3 & 3.72E-05 & 2.97E-05 & 4.89E-02 & 3.71E-02 & 6.70E-01 & 3.26E-03 & 1.22E-03 & 2.67E-01 & 1.14E-01 & 6.36E-02 & 6.54E+00 & 1.63E-01 & 1.02E-01 & 7.17E+00 \\ \hline
    [512]*4 & 3.80E-05 & 3.15E-05 & 4.94E-02 & 3.77E-02 & 9.84E-01 & 3.17E-03 & 1.25E-03 & 1.92E-01 & 1.12E-01 & 6.45E-02 & 4.73E+00 & 1.62E-01 & 1.04E-01 & 5.19E+00 \\ \hline
    \end{tabular}
    \end{adjustbox}
\end{sidewaystable}

\section{Speed-ups with different PINN architectures}
\label{sec:more_speedups}

We performed extensive study on the speed-ups obtained by the trained PINN model inference as compared to the convectional numerical root finding using Python SciPy \texttt{fsolve}. In these experiments, we have used PINN models with neural network having 32, 64, 128, 256, 512 neurons in the hidden layers, and 2, 3 or 4 hidden layers. We have also varied the batch size and inference hardware for each of these NNs. Table \ref{tbl:model_performance_python_V100} and \ref{tbl:model_performance_python_P2000} present speed-ups obtained on the workstation (NVIDIA Tesla V100) and notebook (NVIDIA Qudro P2000) GPUs, respectively. These experiments serve as an extension of the results presented in Table \ref{tbl:Speedups_compared_to_scipy_fsolve}. Readers are refereed to section \ref{sec:time-metrics} for more details on the setup of these experiments.

\begin{table*}
    \caption{\label{tbl:model_performance_python_V100} Model performance on Nvidia Tesla V100 Workstation}
    \centering{}
    \begin{tabular}{|c|c|c|c|c|c|c|c|}
    \hline
    \multirow{2}{*}{Architecture} & \multicolumn{7}{c|}{Batch size} \\
    \cline{2-8} & $10^{0}$ & $10^{1}$ & $10^{2}$ & $10^{3}$ & $10^{4}$ & $10^{5}$ & $10^{6}$ \\ \hline
    [32]*2 & 1.85E+02 & 1.87E+03 & 1.89E+04 & 1.77E+05 & 1.32E+06 & 4.76E+06 & 5.66E+06 \\ \hline
    [32]*3 & 1.73E+02 & 1.79E+03 & 1.77E+04 & 1.80E+05 & 1.33E+06 & 4.86E+06 & 6.27E+06 \\ \hline
    [32]*4 & 1.60E+02 & 1.67E+03 & 1.66E+04 & 1.77E+05 & 1.33E+06 & 4.41E+06 & 4.39E+06 \\ \hline
    [64]*2 & 1.73E+02 & 1.69E+03 & 1.91E+04 & 1.79E+05 & 1.39E+06 & 4.09E+06 & 5.59E+06 \\ \hline
    [64]*3 & 1.82E+02 & 1.86E+03 & 1.92E+04 & 1.77E+05 & 1.33E+06 & 3.45E+06 & 3.20E+06 \\ \hline
    [64]*4 & 1.68E+02 & 1.77E+03 & 1.97E+04 & 1.72E+05 & 1.28E+06 & 2.99E+06 & 2.24E+06 \\ \hline
    [128]*2 & 1.83E+02 & 1.91E+03 & 1.86E+04 & 1.76E+05 & 1.22E+06 & 2.88E+06 & 3.17E+06 \\ \hline
    [128]*3 & 1.84E+02 & 1.82E+03 & 1.77E+04 & 1.73E+05 & 1.08E+06 & 2.26E+06 & 2.57E+06 \\ \hline
    [128]*4 & 1.70E+02 & 1.76E+03 & 1.69E+04 & 1.70E+05 & 9.65E+05 & 1.86E+06 & 2.07E+06 \\ \hline
    [256]*2 & 1.94E+02 & 1.79E+03 & 1.77E+04 & 1.74E+05 & 8.88E+05 & 1.66E+06 & 1.72E+06 \\ \hline
    [256]*3 & 1.78E+02 & 1.82E+03 & 1.73E+04 & 1.77E+05 & 7.00E+05 & 1.09E+06 & 1.16E+06 \\ \hline
    [256]*4 & 1.68E+02 & 1.76E+03 & 1.80E+04 & 1.53E+05 & 5.65E+05 & 8.16E+05 & 8.76E+05 \\ \hline
    [512]*2 & 1.81E+02 & 1.82E+03 & 1.85E+04 & 1.61E+05 & 5.12E+05 & 7.02E+05 & 7.37E+05 \\ \hline
    [512]*3 & 1.90E+02 & 1.81E+03 & 1.73E+04 & 1.24E+05 & 3.38E+05 & 3.99E+05 & 4.26E+05 \\ \hline
    [512]*4 & 1.73E+02 & 1.80E+03 & 1.63E+04 & 1.01E+05 & 2.52E+05 & 2.96E+05 & 3.06E+05 \\ \hline
    \end{tabular}
\end{table*}

\begin{table*}
    \caption{\label{tbl:model_performance_python_P2000} Model performance on Nvidia Quadro P2000 Notebook. OOM = Out Of Memory}
    \centering{}
    \begin{tabular}{|c|c|c|c|c|c|c|c|}
    \hline
    \multirow{2}{*}{Architecture} & \multicolumn{7}{c|}{Batch size} \\
    \cline{2-8} & $10^{0}$ & $10^{1}$ & $10^{2}$ & $10^{3}$ & $10^{4}$ & $10^{5}$ & $10^{6}$ \\ \hline
    [32]*2 & 9.13E+01 & 9.09E+02 & 9.02E+03 & 7.40E+04 & 4.33E+05 & 6.75E+05 & 7.34E+05 \\ \hline
    [32]*3 & 9.59E+01 & 1.01E+03 & 1.01E+04 & 8.79E+04 & 3.19E+05 & 4.59E+05 & 4.80E+05 \\ \hline
    [32]*4 & 8.60E+01 & 8.77E+02 & 9.33E+03 & 8.75E+04 & 2.42E+05 & 3.47E+05 & 3.57E+05 \\ \hline
    [64]*2 & 9.42E+01 & 9.77E+02 & 9.01E+03 & 9.69E+04 & 3.23E+05 & 4.37E+05 & 4.56E+05 \\ \hline
    [64]*3 & 8.46E+01 & 9.34E+02 & 9.85E+03 & 7.38E+04 & 2.16E+05 & 2.66E+05 & 2.76E+05 \\ \hline
    [64]*4 & 6.28E+01 & 4.79E+02 & 4.69E+03 & 4.95E+04 & 1.47E+05 & 1.90E+05 & 1.97E+05 \\ \hline
    [128]*2 & 8.12E+01 & 8.50E+02 & 7.96E+03 & 4.74E+04 & 1.27E+05 & 1.51E+05 & 1.52E+05 \\ \hline
    [128]*3 & 8.49E+01 & 7.79E+02 & 8.08E+03 & 4.04E+04 & 7.52E+04 & 8.37E+04 & 8.57E+04 \\ \hline
    [128]*4 & 6.75E+01 & 7.51E+02 & 6.46E+03 & 3.20E+04 & 5.45E+04 & 5.83E+04 & 5.94E+04 \\ \hline
    [256]*2 & 8.53E+01 & 8.42E+02 & 7.46E+03 & 2.75E+04 & 4.20E+04 & 4.52E+04 & OOM \\ \hline
    [256]*3 & 7.17E+01 & 7.81E+02 & 5.25E+03 & 1.51E+04 & 2.29E+04 & 2.37E+04 & OOM \\ \hline
    [256]*4 & 6.11E+01 & 7.26E+02 & 4.36E+03 & 1.11E+04 & 1.52E+04 & 1.61E+04 & OOM \\ \hline
    [512]*2 & 8.76E+01 & 8.35E+02 & 3.38E+03 & 9.64E+03 & 1.22E+04 & 1.24E+04 & OOM \\ \hline
    [512]*3 & 7.00E+01 & 7.89E+02 & 2.76E+03 & 5.32E+03 & 6.26E+03 & 6.41E+03 & OOM \\ \hline
    [512]*4 & 5.66E+01 & 6.42E+02 & 2.03E+03 & 3.56E+03 & 4.20E+03 & 4.23E+03 & OOM \\ \hline
    \end{tabular}
\end{table*}
\newpage

\newpage
\bibliographystyle{elsarticle-num} 
\bibliography{cas-refs}
\end{document}